\documentclass[runningheads]{llncs}
\usepackage[T1]{fontenc}
\usepackage{graphicx}
\usepackage{amsmath,amssymb}
\usepackage{bm}
\usepackage{algorithm}
\usepackage{algpseudocode}

\graphicspath{{./figures/}}
\makeatletter
\DeclareMathOperator\softplus{sfp}
\DeclareMathOperator\sigmoid{sig}
\newcommand{\argmin}{\mathop{\rm arg~min}\limits}
\newcommand{\figcaption}[1]{\def\@captype{figure}\caption{#1}}
\newcommand{\tblcaption}[1]{\def\@captype{table}\caption{#1}}
\newcommand{\Hline}{\noalign{\hrule height 1pt}} 
\makeatother

\begin{document}
\title{Improving Interpretability of Scores in Anomaly Detection Based on Gaussian--Bernoulli Restricted Boltzmann Machine}
\titlerunning{Improving Interpretability of Scores in AD Based on GBRBM}
%
%
\author{Kaiji Sekimoto\inst{1}\orcidID{0000-0003-1359-6010} \and
Muneki Yasuda\inst{1}\orcidID{0000-0001-5531-9842}}
\authorrunning{K. Sekimoto and M. Yasuda}
%
\institute{Graduate School of Science and Engineering, Yamagata University 4-3-16 Jonan, Yonezawa, Yamagata 992-8510, Japan\\ \email{k.sekimoto1002@gmail.com, muneki@yz.yamagata-u.ac.jp}}
\maketitle              
\begin{abstract}
Gaussian--Bernoulli restricted Boltzmann machines (GBRBMs) are often used for semi-supervised anomaly detection, where they are trained using only normal data points.
In GBRBM-based anomaly detection, normal and anomalous data are classified based on a score that is identical to an energy function of the marginal GBRBM. 
However, the classification threshold is difficult to set to an appropriate value, as this score cannot be interpreted.
In this study, we propose a measure that improves score's interpretability based on its cumulative distribution, and establish a guideline for setting the threshold using the interpretable measure. 
The results of numerical experiments show that the guideline is reasonable when setting the threshold solely using normal data points.
Moreover, because identifying the measure involves computationally infeasible evaluation of the minimum score value, we also propose an evaluation method for the minimum score based on simulated annealing, which is widely used for optimization problems.
The proposed evaluation method was also validated using numerical experiments.


\end{abstract}
\section{Introduction}

As a machine-learning technology that automatically detects anomalous data points that significantly deviate from the norm, anomaly detection (AD) is utilized within diverse research areas and application domains.
AD systems are constructed to suit the nature of available data points; thus, given a sufficient number of normal and anomalous data points, a supervised AD is used.
Because anomalies are typically rare and may be unknown, AD systems are widely constructed in a semi-supervised manner solely using normal data.
In semi-supervised AD, restricted Boltzmann machines (RBMs)~\cite{smolensky1986,hinton2002training}, which are energy-based models in the form of Markov random fields, are often adopted~\cite{Pumsirirat2018,fiore2013,seo2016improvement}.
Whereas standard RBMs handle binary data points, Gaussian--Bernoulli RBMs (GBRBMs)~\cite{cho2011} can handle continuous data.
RBMs are used for not only AD, but also dimensionality reduction~\cite{hinton2006}, pre-training of deep neural networks \cite{salakhutdinov2010}, imputation of incomplete data points \cite{fissore2019}, representation of many-body quantum states~\cite{Melko2019,PhysRevX.11.031034}, design of solid electrolytes~\cite{HatakeyamaSato2022}, and other applications.

In semi-supervised AD, the trained GBRBM expresses a probability density function of normal data points.
The density value can be used as an AD score, with higher values indicating higher normality.
However, because the density value is computationally difficult to evaluate, the energy function is used as an alternative~\cite{Do2018}.
We henceforth refer to the output of this function as the ``free energy (FE) score.''
Using a predefined threshold of the FE score, data are classified as normal (or anomalous) when their FE scores are lower (or higher) than the threshold.
Although an appropriate threshold must be set to ensure classification accuracy, the threshold cannot be set without some additive measure (e.g., F-score or Mathews correlation coefficient (MCC)~\cite{MATTHEWS1975442}), as we cannot interpret whether a given FE score is high or low.
Consequently, this measure requires sufficient normal and anomaly data points, with the latter being difficult to prepare.

In this paper, we propose a measure that improves the interpretability of the FE score using only normal data points, based upon the cumulative distribution of FE scores for normal data.
When the cumulative distribution returns a high probability (i.e., the FE score is high), the data point is located in the tails of the probability density function for normal data. 
Because the tails of this function represent rare cases, the given data point can then be regarded as anomalous.
Thus,the cumulative distribution value is a probabilistic score that can statistically interpret the degree of anomaly for a given data point.
We can set the threshold using the cumulative distribution: when we specify a desired anomalous probability value, the corresponding FE score is set as the threshold.
However, identifying the cumulative distribution is computationally difficult because it involves infeasible multiple integration.
Instead, our objective was to obtain the cumulative distribution by training a one-dimentional GBRBM.

Identifying the cumulative distribution requires an evaluation of the minimum FE score, which is computationally difficult and therefore the minimum evaluation requires an approximation method.
Our previously proposed method~\cite{sekimoto2023quasi} employs a GBRBM with replicated hidden layers and evaluates the minimum FE score by approximating the partition function of the replicated GBRBM via annealed importance sampling (AIS)~\cite{neal2001annealed}, an importance sampling technique based on simulated annealing (SA)~\cite{Geman1984-wh}.
Although the approximation accuracy increases as the number of replications of the hidden layer increases, this incurs a high computational cost.
Extending upon this approach, we propose an evaluation method that does not require replication, instead using SA to locate data points corresponding to the minimum energy value.
Thus, the proposed method can obtain not only the minimum FE score, but also the corresponding data.

In Section~\ref{sec:GBRBM-based_anomaly_detection}, we define GBRBMs and discuss semi-supervised AD within their context. 
In Section~\ref{sec:minimum_FE_score}, we propose an evaluation method for the minimum FE score, which is needed to obtain a guideline to set the threshold, and present the results of a comparative experiment.
In Section~\ref{sec:Measure_and_Guideline}, we provide a guideline for setting the threshold using a measure that improves the interpretability of the FE score, and verify the method through numerical experiments.
Section~\ref{sec:conclusion_future-works} concludes the paper.

\section{Anomaly Detection Based on Gaussian--Bernoulli Restricted Boltzmann Machine} \label{sec:GBRBM-based_anomaly_detection}

\subsection{Gaussian--Bernoulli Restricted Boltzmann Machine} \label{sec:GBRBM}

GBRBMs are Markov random fields defined on complete bipartite graphs and consisting of visible and hidden layers with continuous visible variables $\bm{v}:=\{v_i\in\mathbb{R}\mid i\in\mathcal{V}\}$ and discrete hidden variables $\bm{h}:=\{h_j\in\{0,1\}\mid j\in\mathcal{H}\}$, respectively.
Here, $\mathcal{V}:=\{1,2,\ldots,n_v\}$ denotes a set of indices of the visible variables and $\mathcal{H}:=\{1,2,\ldots,n_h\}$ denotes that of the hidden variables.
The energy function of the GBRBM is defined by
\begin{align}
E_\theta(\bm{v},\bm{h}) := \sum_{i\in\mathcal{V}}\frac{v_i^2}{2\softplus(\sigma_i)} - \sum_{i\in\mathcal{V}} b_iv_i - \sum_{j\in\mathcal{H}} c_jh_j - \sum_{i\in\mathcal{V}}\sum_{j\in\mathcal{H}} w_{i,j}v_ih_j,
\label{eq:GBRBM_energy_func}
\end{align}
where $\softplus(x):=\ln(1+e^x)$ denotes the softplus function.
Here, $\bm{b}:=\{b_i\in\mathbb{R}\mid i\in\mathcal{V}\}$, $\bm{c}:=\{c_j\in\mathbb{R}\mid j\in\mathcal{H}\}$, $\bm{w}:=\{w_{i,j}\in\mathbb{R}\mid i\in\mathcal{V}, j\in\mathcal{H}\}$, and $\bm{\sigma}:=\{\sigma_i\in\mathbb{R}\mid i\in\mathcal{V}\}$ are learning parameters collectively denoted by $\theta:=\bm{b}\cup\bm{c}\cup\bm{w}\cup\bm{\sigma}$.
The energy function in Eq.~\eqref{eq:GBRBM_energy_func} represents a modified version of that of the standard GBRBM~\cite{muneki2023new}.
Using this energy function, the GBRBM is defined by 
\begin{align}
P_\theta(\bm{v},\bm{h}) := \frac{1}{Z_\theta}\exp (-E_\theta(\bm{v},\bm{h})), 
\label{eq:GBRBM}
\end{align}
where $Z_\theta$ denotes the partition function expressed as
\begin{align}
Z_\theta := \int d\bm{v}\sum_{\bm{h}}\exp (-E_\theta(\bm{v},\bm{h})), 
\label{eq:GBRBM_partition_func}
\end{align}
where $\int d\bm{v}$ is a multiple integration over $\bm{v}\in\mathbb{R}^{n_v}$ and $\sum_{\bm{h}}$ is a multiple summation over $\bm{h}\in\{0,1\}^{n_h}$.
The visible variables correspond to a data point, whereas, the hidden variables control the expression power of the GBRBM, which increases with $n_h$.

The conditional distribution of visible variables given the hidden variables is expressed as
\begin{align}
P_\theta(\bm{v}\mid\bm{h}) = \prod_{i\in\mathcal{V}}\mathcal{N}(v_i\mid\lambda_i(\bm{h}),\softplus(\sigma_i)),
\label{eq:GBRBM_dist_v|h}
\end{align}
where $\lambda_i(\bm{h}) := \softplus(\sigma_i)(b_i + \sum_{j\in\mathcal{H}} w_{i,j}h_j)$ and $\mathcal{N}(\cdot\mid\mu,\sigma^2)$ denotes the normal distribution with mean $\mu$ and variance $\sigma^2$.
The conditional distribution for hidden variables is expressed as
\begin{align}
P_\theta(\bm{h}\mid\bm{v}) = \prod_{j\in\mathcal{H}} \frac{\exp(\tau_j(\bm{v})h_j)}{1 + \exp(\tau_j(\bm{v}))} = \prod_{j\in\mathcal{H}} \mathrm{Ber}\bigl[h_j\mid p_j(\bm{v})\bigr],
\label{eq:GBRBM_dist_h|v}
\end{align}
where $\tau_j(\bm{v}) := c_j + \sum_{i\in\mathcal{V}} w_{i,j}v_i$ and $p_j(\bm{v}):=\sigmoid(\tau_j(\bm{v}))$ denotes the probability that $h_{j}=1$ given $\bm{v}$, with $\sigmoid(x):=1/(1+e^{-x})$ being the sigmoid function.
Here, $\mathrm{Ber}(\cdot\mid p)$ denotes the Bernoulli distribution with a probability $p$ that the assigned binary variable takes 1.
From Eqs.~\eqref{eq:GBRBM_dist_v|h} and \eqref{eq:GBRBM_dist_h|v}, the variables in one layer are mutually independent when those in the other layer are given.
This property is known as conditional independence and facilitates the blocked Gibbs sampling~\cite{bishop2006pattern}, which is an efficient Markov chain Monte Carlo (MCMC) method.
The computational cost of the blocked Gibbs sampling on the GBRBM is proportional to $n_v n_h$.

\subsection{Semi-Supervised Anomaly Detection} \label{sec:semi-supervised_AD}

Suppose that we obtain a dataset consisting of $N$ normal data points corresponding to $\bm{v}$, $D:=\{\mathbf{d}^{(\mu)}\in\mathbb{R}^{n_v}\mid\mu=1,2,\ldots,N\}$.
The GBRBM is trained by maximizing the log-likelihood function defined by
\begin{align}
\ell(\theta) := \frac{1}{N}\sum_{\mu=1}^N \ln P_\theta\bigl(\mathbf{d}^{(\mu)}\bigr),
\label{eq:log-likelihood}
\end{align}
where $P_\theta(\bm{v})$ is the marginal GBRBM expressed as
\begin{align}
P_\theta(\bm{v}) = \sum_{\bm{h}} P_\theta(\bm{v},\bm{h}) = \frac{1}{Z_\theta}\exp(-f_\theta(\bm{v})).
\label{eq:GBRBM_visible}
\end{align}
Here, $f_\theta(\bm{v})$ is the energy function of the marginal distribution defined by
\begin{align}
f_\theta(\bm{v}) := -\ln \sum_{\bm{h}}\exp(-E_\theta(\bm{v},\bm{h})). 
\label{eq:GBRBM_energy_v}
\end{align}
Although the energy function in Eq.~\eqref{eq:GBRBM_energy_v} is referred to as free energy in AD~\cite{Do2018}, (true) free energy is originally defined by $F_\theta := - \ln Z_\theta$.
To clearly distinguish between the two, we refer to the energy function in Eq.~\eqref{eq:GBRBM_energy_v} as ``the FE score.''
Equation.~\eqref{eq:log-likelihood} is maximized by a gradient ascent method using gradients with respect to $\theta$, which include expectations on the GBRBM.
Because these expectations involve intractable multiple integrations and summations, the maximization of Eq.~\eqref{eq:log-likelihood} is computationally difficult; hence, the expectation is approximated.
As mentioned in Section~\ref{sec:GBRBM}, because sampling on GBRBMs can be conducted efficiently, it is appropriate to use a sampling-based approximation method such as the contrastive divergence method~\cite{hinton2002training}, parallel tempering~\cite{desjardins2010parallel}, or spatial Monte Carlo integration~\cite{sekimoto2023effective}.

After training, the marginal GBRBM in Eq.~\eqref{eq:GBRBM_visible} expresses a probability density of the normal data points.
Therefore, the density value can be considered a score of AD, where higher density values are associated with higher normality. 
However, the partition function contained in the marginal GBRBM is computationally difficult to evaluate for the same reason as the expectation.
From Eq.~\eqref{eq:GBRBM_visible}, the marginal GBRBM has a monotonic inverse relationship with the FE score.
Because the FE score is feasible to evaluate, it can be used to substitute for the density value as the AD score.
For a predefined threshold $\kappa\in\mathbb{R}$, a given data point $\mathbf{d}\in\mathbb{R}^{n_v}$ is classified as normal when $f_\theta(\mathbf{d}) < \kappa$ and anomalous when $f_\theta(\mathbf{d}) > \kappa$.
Selecting a proper threshold is critical for an accurate classification.
However, the proper threshold cannot be determined using only normal data points, as the FE score is difficult to interpret.
We therefore propose a measure that can be used to interpret the FE score and provide a guideline for setting the threshold using only normal data.

\section{Evaluation of Minimum Free Energy Score} \label{sec:minimum_FE_score}

To interpret the FE score, we must first evaluate the minimum FE score defined by
\begin{align}
\mathrm{f}^* := \min_{\bm{v}} f_\theta(\bm{v}).
\label{eq:Minimum_FE_score}
\end{align}
This minimization is computationally infeasible because it must search the entire $n_v$-dimensional space.
We therefore propose an evaluation method for $\mathrm{f}^*$ based on SA, and numerically investigate it through a comparative experiment.

\subsection{Proposed Method Based on Simulated Annealing} \label{sec:Proposed_SA}

\begin{figure}[t]
\centering
\includegraphics[width=0.95\linewidth]{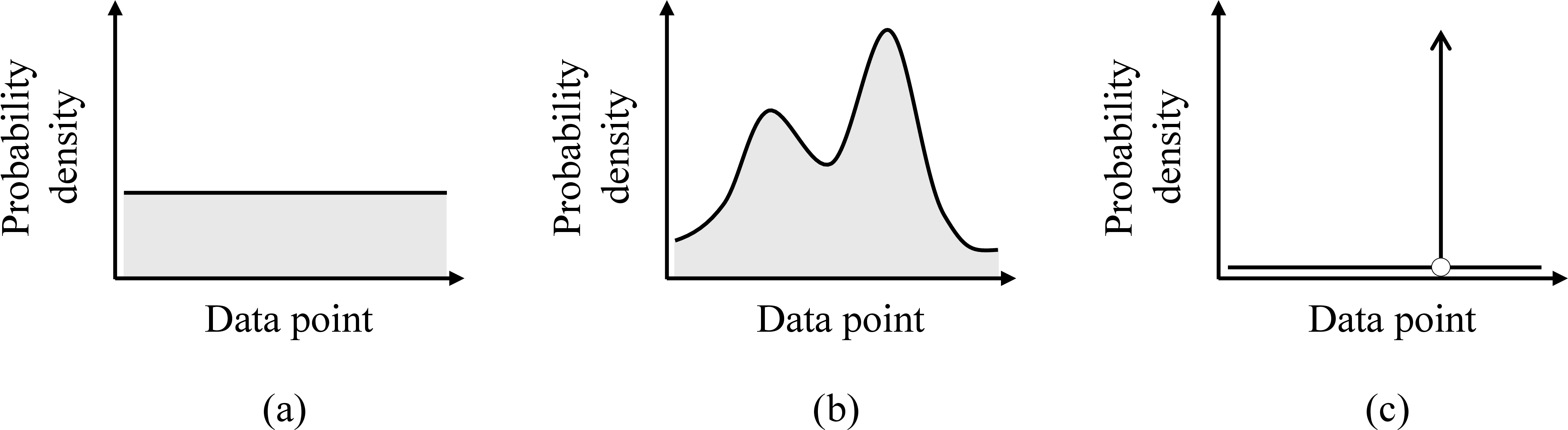}
\caption{
Illustration of extended marginal GBRBM in Eq.~\eqref{eq:extended_GBRBM} against various $\beta$.
$\beta$ manages the magnitude of the density ratio among data points; when (a) $\beta\rightarrow 0$, (b) $\beta=1$, and (c) $\beta\rightarrow\infty$, the extended marginal GBRBM becomes a uniform distribution, the marginal GBRBM in Eq.~\eqref{eq:GBRBM_visible}, and the delta function $\delta(f_\theta(\bm{v}) - \mathrm{f}^*)$, respectively.
}
\label{fig:inverse_temp}
\end{figure}

By introducing the inverse temperature $\beta > 0$, the marginal GBRBM in Eq.~\eqref{eq:GBRBM_visible} is extended to
\begin{align}
P_\theta(\bm{v}\mid\beta) := \frac{1}{Z_\theta(\beta)}\exp(-\beta f_\theta(\bm{v})),
\label{eq:extended_GBRBM}
\end{align}
where $Z_\theta(\beta) := \int d\bm{v} \exp(-\beta f_\theta(\bm{v}))$ denotes the partition function of the extended marginal GBRBM.
When the FE score of a data point $\mathbf{v}_1$ exceeds that of another data point $\mathbf{v}_2$, the FE-score ratio between the two points, $\exp(-\beta f_\theta(\mathbf{v}_1))/\exp(-\beta f_\theta(\mathbf{v}_2))$, is positive and increases with $\beta$.
Therefore, the inverse temperature controls the complexity of the extended marginal GBRBM in Eq.~\eqref{eq:extended_GBRBM} (see Fig.~\ref{fig:inverse_temp}).
Based upon SA, the proposed method searches $\mathrm{f}^*$ by performing an MCMC over $P(\bm{v}\mid\beta)$ while gradually increasing the inverse temperature.
As an efficient MCMC method for GBRBMs, we adopted blocked Gibbs sampling for the proposed method.
To implement this method, the extended marginal GBRBM in Eq.~\eqref{eq:extended_GBRBM} is reverted to the GBRBM form in Eq.~\eqref{eq:GBRBM} by the following ingenuity.

\begin{figure}[t]
\centering
\includegraphics[width=0.5\linewidth]{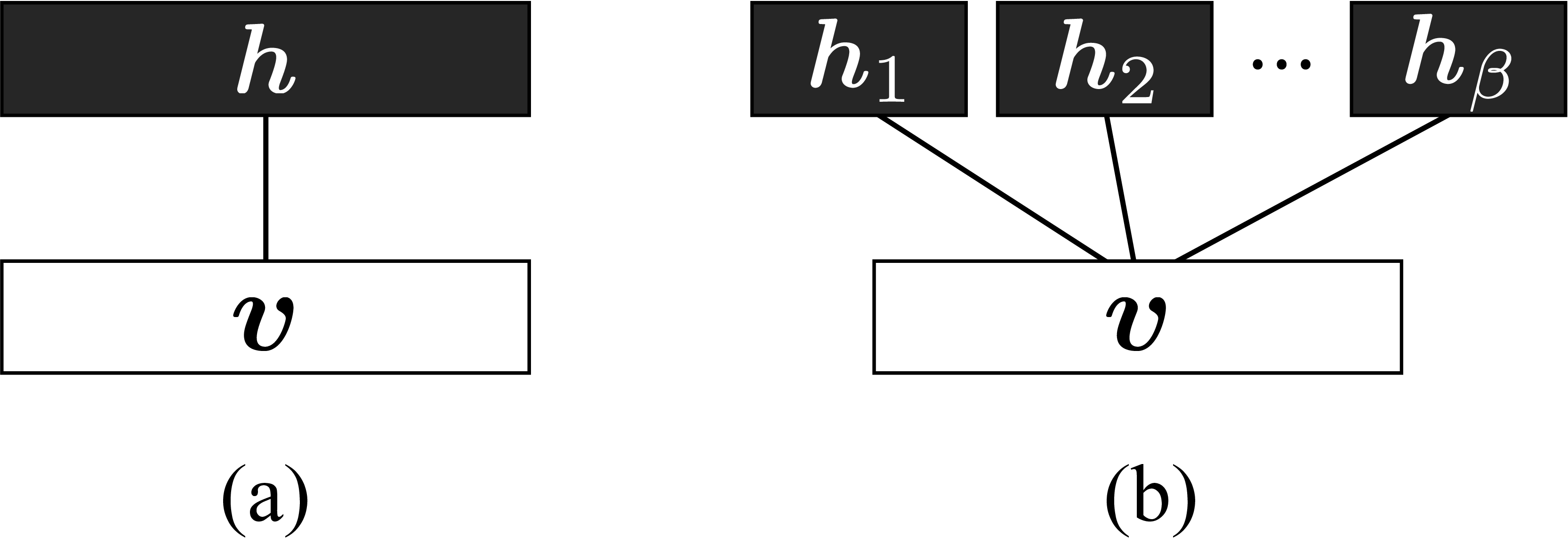}
\caption{
Illustration of (a) original GBRBM in Eq.~\eqref{eq:GBRBM} and (b) replicated GBRBM in Eq.~\eqref{eq:replicated_GBRBM}.
}
\label{fig:replicated_GBRBM}
\end{figure}

Suppose that the inverse temperature is a natural number, i.e, $\beta\in\mathbb{N}$.
In this case, the extended marginal GBRBM in Eq.~\eqref{eq:extended_GBRBM} is rewritten as
\begin{align*}
P_{\theta}(\bm{v} \mid \beta) = \frac{1}{Z_{\theta}(\beta)}\sum_{\bm{h}_1}\sum_{\bm{h}_2}\cdots \sum_{\bm{h}_{\beta}} 
\exp\Biggl( -\sum_{r=1}^{\beta} E_{\theta}(\bm{v}, \bm{h}_r)\Biggr),
\end{align*}
where $\bm{h}_r := \{h_{r,j}\in\{0,1\}\mid j\in\mathcal{H}\}$ denotes the $r$-th replica of the hidden layer, and $h_{r,j}$ denotes the $r$-th replica of $h_j$.
From the above equation, the extended marginal GBRBM can be reverted to that with $\beta$-replicated hidden layers (see Fig.~\ref{fig:replicated_GBRBM}) defined by
\begin{align}
P_{\theta}(\bm{v},\bm{H} \mid \beta) := \frac{1}{Z_{\theta}(\beta)}
\exp\Biggl( -\sum_{r=1}^{\beta} E_{\theta}(\bm{v}, \bm{h}_r)\Biggr),
\label{eq:replicated_GBRBM}
\end{align}
where $\bm{H} := \{\bm{h}_r\mid r = 1,2,\ldots,\beta\}$ denotes a set of replicated hidden variables.
From the saddle point method, the partition function $Z_\theta(\beta)$ asymptotically approaches to $\exp(-\beta \mathrm{f}^*)$ as $\beta$ increases; that is, the relation,
\begin{align*}
\mathrm{f}^* = \lim_{\beta\rightarrow\infty} -\frac{1}{\beta}\ln Z_\theta(\beta),
\end{align*}
holds.
In our previous method~\cite{sekimoto2023quasi}, the minimum FE score is evaluated by approximating $Z_{\theta}(\beta)$ using AIS~\cite{neal2001annealed}. 
The computational cost of the previous method is proportional to the number of replicas (i.e., the inverse temperature) $\beta$.
Consequently, a sufficiently high $\beta$ to obtain an accurate approximation incurs a considerable cost.

We consider the conditional distribution for the replicated hidden variables $\bm{H}$ expressed as
\begin{align*}
P_\theta(\bm{H}\mid\bm{v},\beta) = \prod_{j\in\mathcal{H}}\prod_{r=1}^\beta \mathrm{Ber}\bigl[h_{r,j}\mid p_j(\bm{v})\bigr].
\end{align*}
In this equation, $h_{r,j}$ for $r=1,2,\ldots,\beta$ are independent and follow the same Bernoulli distribution with a mean of $p_j(\bm{v})$ and variance of $p_j(\bm{v})(1-p_j(\bm{v}))$.
Therefore, from the central limit theorem, the distribution of the sample average $\hat{h}_{j} := \beta^{-1}\sum_{r=1}^\beta h_{r,j}$ converges to a normal distribution with mean $p_j(\bm{v})$ and variance $\beta^{-1}p_j(\bm{v})(1-p_j(\bm{v}))$ when $\beta$ is sufficiently large.
That is, the asymptotic distribution of $\hat{\bm{h}}:=\{\hat{h}_j\mid j\in\mathcal{H}\}$ is defined by
\begin{align}
Q_\theta(\hat{\bm{h}}\mid\bm{v},\beta) :=\prod_{j\in\mathcal{H}}\mathcal{N}\bigl[\hat{h}_j\mid p_j(\bm{v}),\beta^{-1} p_j(\bm{v})(1-p_j(\bm{v}))\bigr].
\label{eq:asymptDist_h|v}
\end{align}
Note that we henceforth treat $\hat{h}_j$ as a continuous random variable, i.e., $\hat{h}_j\in\mathbb{R}$.
Furthermore, the conditional distribution of $\bm{v}$ given $\hat{\bm{h}}$ is expressed as
\begin{align}
P_\theta(\bm{v}\mid\hat{\bm{h}},\beta) = \prod_{i\in\mathcal{V}}\mathcal{N}(v_i\mid\lambda_i(\hat{\bm{h}}),\beta^{-1}\softplus(\sigma_i)).
\label{eq:replica_GBRBM_v|h}
\end{align}
The computational cost of the blocked Gibbs sampling based on the conditional distributions in Eqs.~\eqref{eq:asymptDist_h|v} and \eqref{eq:replica_GBRBM_v|h} does not depend on the value of $\beta$.
As $\beta$ increases, a sample point for $\bm{v}$ generated by this sampling asymptotically follows the replicated GBRBM in Eq.~\eqref{eq:replicated_GBRBM}, i.e., the extended marginal GBRBM in Eq.~\eqref{eq:extended_GBRBM}.
In the limit of $\beta\rightarrow\infty$, the sample point can be interpreted as drawn from the extended marginal GBRBM shown in Fig. \ref{fig:inverse_temp}(c).
Based on this sampling, the proposed method yields the minimum FE score and corresponding data point expressed as
\begin{align}
\mathbf{v}^* := \argmin_{\bm{v}} f_\theta(\bm{v}).
\label{eq:MIN_ENERGY_data_point}
\end{align}

We prepared a sequence of inverse temperatures expressed as $\beta_1 < \beta_2 < \cdots < \beta_K$, where $K$ is the size of the annealing schedule and $\beta_K$ is set as $\beta_K\rightarrow\infty$ from the above discussion.
Although the inverse temperature was assumed to be a natural number in order to construct the replicated GBRBM, we handled it as a real number for simplicity.
For each inverse temperature $\beta_k$, the transition probability can be modeled as
\begin{align*}
T_k(\bm{v}'\mid\bm{v}) = \sum_{\hat{\bm{h}}}P_\theta(\bm{v}'\mid\hat{\bm{h}},\beta_k)Q_\theta(\hat{\bm{h}}\mid\bm{v},\beta_k).
\end{align*}
This transition probability represents a collapsed Gibbs sampling~\cite{yasuda2022free}.
In practice, it can be approximated by the blocked Gibbs sampling to reduce the computational cost. 
Using the sequence of transition probabilities, the minimum FE score $\mathrm{f}^*$ is searched by SA (see Algorithm~\ref{alg:proposed_method_minimum}).
Following $M$ iterations of the SA-based algorithm, we designate the smallest value among the $M$ candidate FE scores as $\mathrm{f}^*$.
From the theory of SA, if $\beta_1$ is set to a tiny positive value, $\beta_K\rightarrow\infty$, and the difference of inverse temperatures $\beta_{k+1}-\beta_k$ is extremely small, then it is theoretically guaranteed that the minimum FE score can be obtained~\cite{Geman1984-wh}.

Because AIS is based upon SA, our previous method employs an algorithm similar to the proposed method, and the magnitude relationship between their computational costs is determined by the sampling cost.
In the previous method, which performs the blocked Gibbs sampling on the replicated GBRBM in Eq.~\eqref{eq:replicated_GBRBM}, the sampling cost is proportional to $n_v n_h \beta$.
By contrast, the sampling cost in the proposed method is proportional to $n_v n_h$ ensuring faster performance.
Moreover, unlike the previous method, our new method obtains the data point corresponding to the FE score.

\begin{algorithm}[t]
\caption{Evaluating minimum FE score and corresponding data point}
\label{alg:proposed_method_minimum}
\begin{algorithmic}
\Require Sequence of inverse temperatures $\{
\beta_k\}$, Initial state $\mathbf{v}_0$, Number of steps $S$
\For{$k=1,\ldots,K$}
\State $\mathbf{v}_k^{(0)} = \mathbf{v}_{k-1}$
\For{$s=1,\ldots,S$} \Comment{$S$-step Blocked Gibbs sampling}
\State $\hat{\mathbf{h}}_k^{(s)}\sim Q_\theta(\hat{\bm{h}}\mid\mathbf{v}_k^{(s-1)},\beta_k)$
\State $\mathbf{v}_k^{(s)}\sim P_\theta(\bm{v}\mid\hat{\mathbf{h}}_k^{(s)},\beta_k)$
\EndFor
\EndFor
\State \Return $f_\theta\bigl(\mathbf{v}_K^{(S)}\bigr)$, $\mathbf{v}_K^{(S)}$ \Comment{Minimum FE score and corresponding data point}
\end{algorithmic} 
\end{algorithm}

\subsection{Comparative Numerical Experiment} \label{sec:conventional_vs_proposed}

We compared our previous and current methods through numerical experiments using GBRBMs trained with a toy dataset and a real dataset.

\subsubsection{Experiment for Toy Dataset} \label{sec:toy_data}

The toy dataset consists of $\{-1,1\}$-binary images of $28\times28$ pixels.
We considered four basic images: (1) all pixels set to $-1$, (2) all pixels set to $1$, (3) the top half is set to $-1$ and the bottom half is set to $1$, and (4) the top half is set to $1$ and the bottom half is set to $-1$.
We designated the basic images from (1) to (3) as normal data point, with all other images regarded as anomalous.
The data were augmented by introducing noise, which controls the distance between basic images, to the pixels of each basic image independently, as well as truncating values greater than $1$ and less than $-1$ to $1$ and $-1$, respectively.
We introduced a Gaussian noise with mean 0 and variance $0.5^2$, and a noise that flips with probability $0.1$.

\begin{figure*}[t]
\centering
\begin{minipage}[h]{0.4\linewidth}
    \centering
    \includegraphics[width=0.9\linewidth]{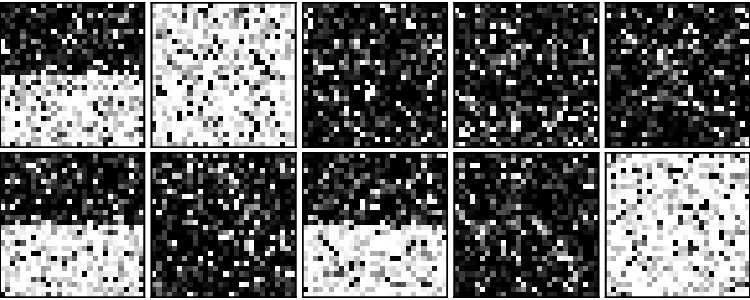} \\
    \figcaption{Random samples from training data of the toy dataset. Black and white pixels indicate values of $-1$ and $+1$, respectively.}
    \label{fig:Toy_Data}
\end{minipage}
\hspace{0.04\linewidth}
\begin{minipage}[h]{0.48\linewidth}
    \centering
    \vspace{-5mm}
    \tblcaption{Estimators of $\mathrm{f}^*$ in Eq.~\eqref{eq:Minimum_FE_score} obtained from previous and proposed methods. The values are the average and standard deviation obtained over 100 experiments.}
    \vspace{3mm}
	\begin{tabular}{l c c}
        \Hline
        && Estimator of $\mathrm{f}^*$ \\ \hline
		Previous~\cite{sekimoto2023quasi} && $-592.40\pm 2.79$ \\ 
		Proposed && $-743.32\pm 2.26\times10^{-5}$ \\ \Hline
	\end{tabular}
    \label{table:result_ToyData}
\end{minipage}
\end{figure*}

Training data were prepared by generating $2,000$ images corresponding to each normal basic image and subsequently introducing noise (see Fig.~\ref{fig:Toy_Data}).
Thus, the training data size was $N=6000$.
We introduced a training GBRBM with $n_v = 784$ and $n_h=500$.
The bias parameters $\bm{b},\bm{c}$ were initialized to 0, the connect parameters $\bm{w}$ were initialized by the Gaussian-type Xavier's initialization~\cite{Xavier2010understanding}, and the variance parameters $\bm{\sigma}$ were initialized by $\ln(e-1)$ so that $\softplus(\sigma_i) = 1$.
To train the model, we used the stochastic gradient method with a batch size of $128$, $1,000$ epochs, and a learning rate optimized by AdaMax~\cite{Adam2015}.
We approximated the GBRBM expectation by the first-order spatial Monte Carlo method~\cite{sekimoto2023effective} using $128$ sample points generated by the $100$-step blocked Gibbs sampling starting from a random state drawn from the multivariate standard normal distribution.
After training, the minimum FE score in Eq.~\eqref{eq:Minimum_FE_score} was evaluated by both methods.
For our previous method, we used a replicated GBRBM with $\beta=20$, an annealing schedule set to $\beta_k=\ln(k)/\ln(K)$ with $K=1000$, initial distribution set to a multivariate standard normal distribution, sample size was fixed to 100, and the 10-step blocked Gibbs sampling used as a transition probability.
In our new method, the annealing schedule was set to $\beta_k^{-1}=\ln(K/k)/\ln(K)$ with $K=1000$, the number of SA iterations was fixed to $M=100$, and the number of steps was $S=10$, with the initial state for each iteration randomly selected from the training data.
The results listed in Table~\ref{table:result_ToyData} demonstrate that our new method successfully yields lower minimum FE scores.

\subsubsection{Experiment for Real Dataset}

\begin{figure}[t]
    \centering
    \includegraphics[width=0.75\linewidth]{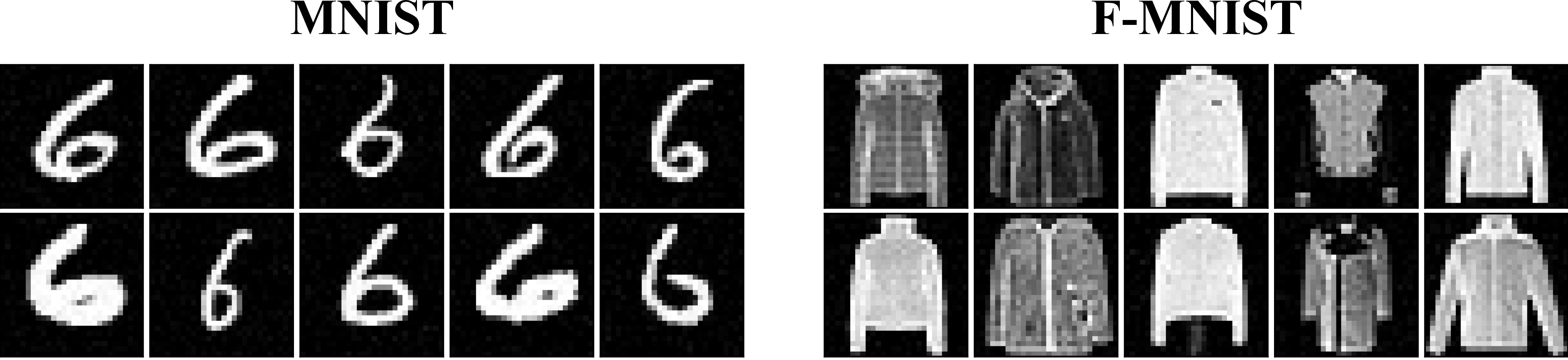} \\
    \figcaption{Random samples from training data obtained from MNIST and F-MNIST.}
    \label{fig:Real_Data}
\end{figure}

We conducted an additional experiment with datasets obtained from MNIST and Fasion MNIST (F-MNIST).
MNIST is a dataset of ten handwritten digit images from 0 to 9, and F-MNIST is a dataset of ten fashion product images.
Both datasets consist of $28\times28$ grayscale images, encompassing $60,000$ training and $10,000$ test data points.
We used images of the digit 6 in MNIST and with the coat item in F-MNIST as normal data points, with all other data regarded as anomalous.
The training data sizes obtained from MNIST and F-MNIST were $N=5918$ and $N=6000$, respectively.
The data points, $\mathbf{d}\in\{0,1,\ldots,255\}^{n_v}$, were normalized by $\mathrm{d}_i\leftarrow 2(\mathrm{d}_i/255)-1$, i.e, $\mathrm{d}_i\in[-1,+1]$; and after normalization, Gaussian noise with mean $0$ and variance $0.05^2$ was introduced to the training data.
Figure~\ref{fig:Real_Data} shows several training data points for MNIST and F-MNIST.
We deployed a training GBRBM with $n_v = 784$ and $n_h=1000$.
The initialization of the learning parameters and the settings of the training and of the previous and proposed methods were identical to those used in the previous experiment.

The results listed in Table~\ref{table:result_RealData} demonstrate that our new method successfully identifies lower FE scores than the previous method.
It is noteworthy that for both datasets, the FE scores obtained by the proposed method are lower than the smallest FE scores for the training data points defined by $\min_{\mu}f_\theta(\mathbf{d}^{(\mu)})$.

\begin{table}[t]
    \centering
    \caption{Estimators of $\mathrm{f}^*$ in Eq.~\eqref{eq:Minimum_FE_score} obtained from the previous and proposed methods. The values are the average and standard deviation obtained over 50 experiments.}
	\begin{tabular}{l c c c c}
        \Hline
        && \multicolumn{3}{c}{Estimator of $\mathrm{f}^*$} \\ \cline{3-5}
        && MNIST && F-MNIST \\ \hline        
		Previous~\cite{sekimoto2023quasi} && $-68051.97\pm 318.09$ && $-26717.63\pm 195.69$ \\ 
		Proposed && $-81484.24\pm 6.99\times10^{-11}$ && $-31803.20\pm 2.30\times10^{-7}$ \\ \Hline
	\end{tabular}
    \label{table:result_RealData}
\end{table}


\section{Interpretation Improvement of Free Energy Score and Guideline for Setting Threshold} \label{sec:Measure_and_Guideline}

\begin{figure}[t]
\centering
\includegraphics[width=\linewidth]{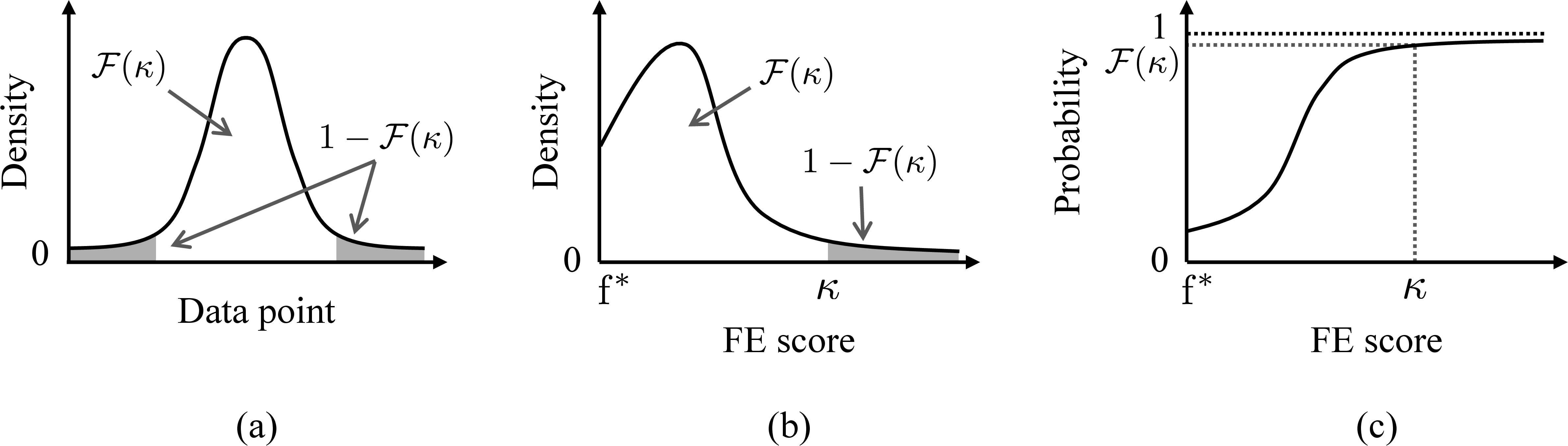}
\caption{
Illustration of (a) the marginal GBRBM in Eq.~\eqref{eq:GBRBM_visible}, (b) the probability density of the FE score in Eq.~\eqref{eq:density_FE_score}, and (c) its cumulative distribution in Eq.~\eqref{eq:cumul_FE_score}.
The gray regions correspond to anomalies when the FE score threshold is set to $\kappa$.
}
\label{fig:proposed_guideline}
\end{figure}

We now propose a measure that improves the interpretability of the FE score, along with a guideline for setting the threshold of the FE score.
Using the trained GBRBM $P_\theta(\bm{v},\bm{h})$, let us consider the probability density of the FE score defined by
\begin{align}
\mathcal{P}(f) := \int d\bm{v}\, \delta(f - f_\theta(\bm{v})) P_\theta(\bm{v})\qquad(\mathrm{f}^*\le f<+\infty),
\label{eq:density_FE_score}
\end{align}
where $\delta(\cdot)$ denotes the Dirac delta function.
Here, the minimum FE score $\mathrm{f}^*$ is obtained from the evaluation method proposed in Section~\ref{sec:minimum_FE_score}.
The proposed measure is designed using the cumulative distribution of the FE score defined by
\begin{align}
\mathcal{F}(f) := \int_{\mathrm{f}^*}^f dx\,\mathcal{P}(x) = \int d\bm{v}\, 1_{[\mathrm{f}^*,f]}(f_\theta(\bm{v})) P_\theta(\bm{v}),
\label{eq:cumul_FE_score}
\end{align}
where $1_A(x)$ denotes the indicator function, which returns 1 if $x\in A$ and 0 if $x\in A$.
This cumulative distribution has a positive monotonic relationship with the FE score; for a given data point $\mathbf{d}\in\mathbb{R}^{n_v}$, when the FE score $f_\theta(\mathbf{d})$ is a high (or low) value, the proposed measure $\mathcal{F}(f_\theta(\mathbf{d}))$ provides a high (or low) probability value (see Fig.~\ref{fig:proposed_guideline}).
Therefore, data points associated with high probability values can be regarded as anomalous.
This means that the proposed measure is a probabilistic score that can be used to statistically interpret the degree of anomaly for given data.
Using the proposed measure, we can formulate a guideline for setting the threshold of the FE score; when we set a predetermined anomalous probability $p_\mathrm{anom}$, the threshold $\kappa$ must be set such that $\mathcal{F}(\kappa)=p_\mathrm{anom}$.
Because the cumulative distribution function is monotonically increasing, the threshold can be set quickly using a binary search.
We note that this guideline is constructed using only normal data.

However, the FE-score density in Eq.~\eqref{eq:density_FE_score} is computationally difficult to calculate, as it involves the intractable multiple integration $\int d\bm{v}$.
Therefore, an evaluation of the proposed measure requires an approximation.
We suppose that the training dataset $D$ was generated from the trained GBRBM and approximate the FE-score density by
\begin{align}
\mathcal{P}(e) \approx \frac{1}{N}\sum_{\mu=1}^N \delta\bigl(e - f_\theta(\mathbf{d}^{(\mu)})\bigr).
\label{eq:empirical_dist_Energy}
\end{align}
This represents the empirical distribution of FE score data points $\{f_\theta(\mathbf{d}^{(\mu)})\mid\mu=1,2,\ldots,N\}$.
We reconstruct $\mathcal{P}(f)$ by the maximum likelihood estimation using a GBRBM with $n_v = 1$, $P_{\bar{\theta}}(v,\bm{h})$.
Here, we recommend standardization or the normalization of the score data as a preprocessing step, as the FE scores typically take negative large values.
We denote the expectation over the GBRBM by $\bar{\mathbb{E}}[\cdots] := \int_{\mathrm{f}^*}^{+\infty}dv \sum_{\bm{h}} (\cdots) P_{\bar{\theta}}(v,\bm{h})$.
The gradients with respect to the learning parameters involves expectations $\bar{\mathbb{E}}[v],\bar{\mathbb{E}}[h_j],\bar{\mathbb{E}}[vh_j]$, and $\bar{\mathbb{E}}[v^2]$.
The expectations, $\bar{\mathbb{E}}[h_j]$ and $\bar{\mathbb{E}}[vh_j]$, are analytically marginalized for the hidden variable and can be rewritten as the expectations for the one-dimensional visible variable:
\begin{align*}
\bar{\mathbb{E}}[h_j] &= \int_{-\infty}^{+\infty}dv \,\,\sigmoid(\tau_j(v)) P_{\bar{\theta}}(v), \\
\bar{\mathbb{E}}[vh_j] &= \int_{-\infty}^{+\infty}dv \,\,v\sigmoid(\tau_j(v)) P_{\bar{\theta}}(v).
\end{align*}
Therefore, the GBRBM can be trained using numerical integration.
Using the trained GBRBM $P_{\bar{\theta}}(v,\bm{h})$, the proposed measure in Eq.~\eqref{eq:cumul_FE_score} is approximated by
\begin{align}
\mathcal{F}(f) \approx \int_{\mathrm{f}^*}^x dx\,P_{\bar{\theta}}(x).
\end{align}
This approximation measure can be exactly evaluated using numerical integration.

We now attempt to set the threshold based on the provided guideline.
Using the trained GBRBM and corresponding training data, we obtain score data points $\{f_\theta(\mathbf{d}^{(\mu)})\mid\mu=1,2,\ldots,N\}$.
The FE score data are normalized to follow the standard normal distribution (i.e., Z-scoring).
After normalization, we train a GBRBM with $n_v = 1$ and $n_h = 50$ to approximate the FE-score density in Eq.~\eqref{eq:density_FE_score}.
For the GBRBM, an initialization of $\bar{\theta}$ was performed as in Section~\ref{sec:conventional_vs_proposed}.
In the training phase, we used the gradient ascent method with AdaMax and fixed the epoch to $1,000,000$.
In the partition function and expectation, the integration for $v$ was performed using the Gauss-Legendre quadrature. 
Following training, we determined the anomalous probability to be $p_\mathrm{anom}=0.9$ and obtained the threshold $\kappa$ such that $\mathcal{F}(\kappa)=p_\mathrm{anom}$.
To validate this threshold, we evaluated the MCC, which is a measure of when sufficient normal and anomalous test data points are available.
For the toy dataset, the numbers of normal and anomalous test data points were prepared by $6,000$ ($2,000$ images corresponding to each normal basic image) and $6,000$, respectively; for the MNIST and F-MNIST datasets, they were prepared by $1,000$ and $9,000$, respectively.
We subsequently introduced noise using the same setup procedure as that used for the training data.

\begin{figure}[t]
\centering
\begin{tabular}{rl}
\begin{minipage}[t]{0.48\linewidth}
\centering
\includegraphics[width=0.9\linewidth]{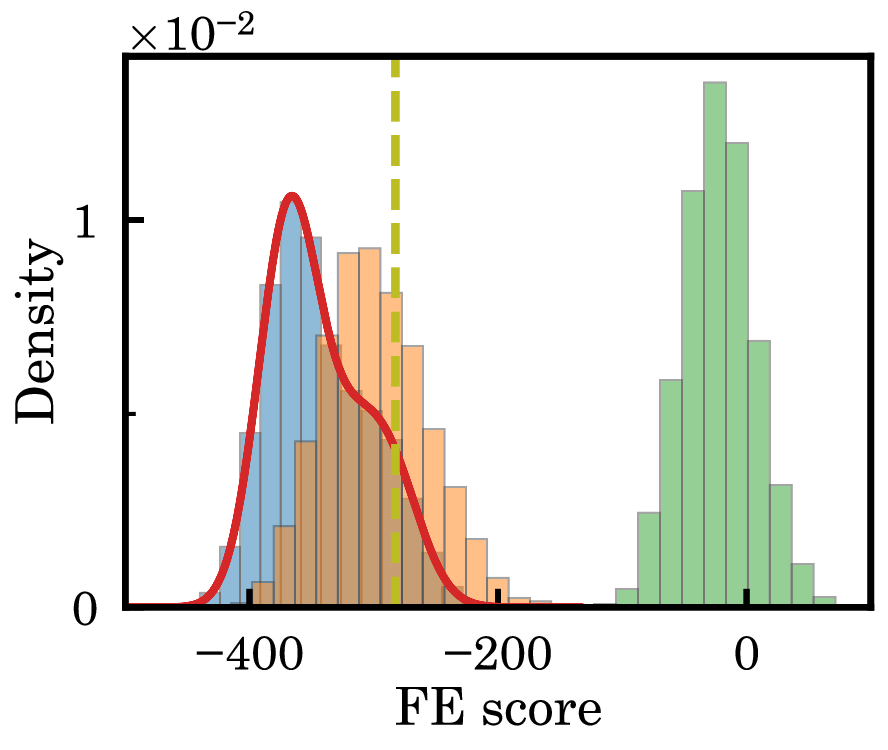} \\
\hspace{0.6cm} (a)
\end{minipage}
\begin{minipage}[t]{0.48\linewidth}
\centering
\includegraphics[width=0.9\linewidth]{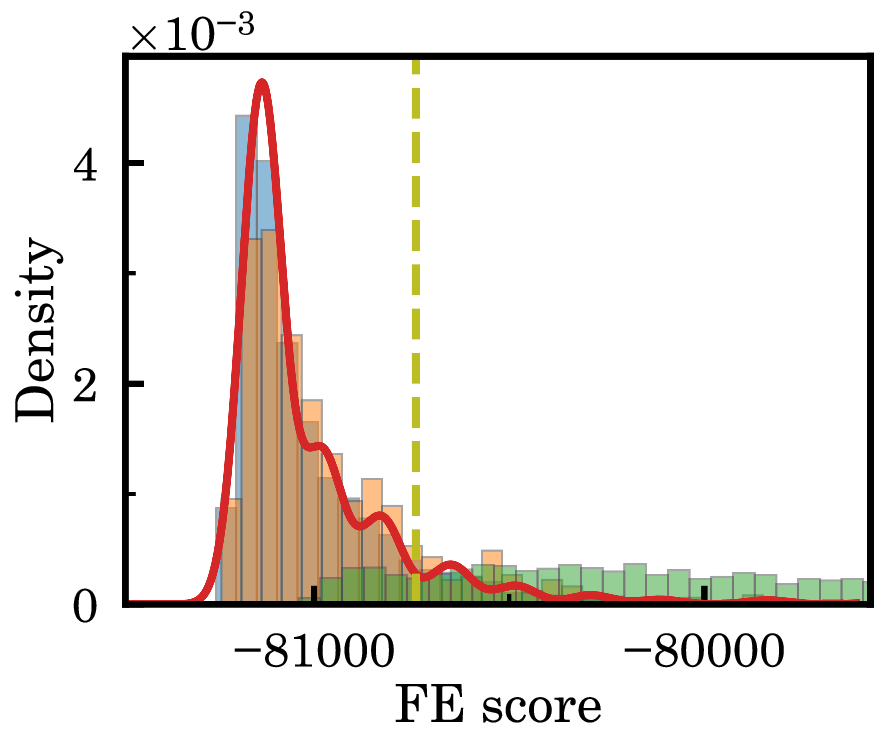} \\
\hspace{0.6cm} (b)
\end{minipage}
\end{tabular}
\ \\
\ \\
\ \\
\begin{tabular}{rl}
\begin{minipage}[t]{0.48\linewidth}
\centering
\includegraphics[width=0.9\linewidth]{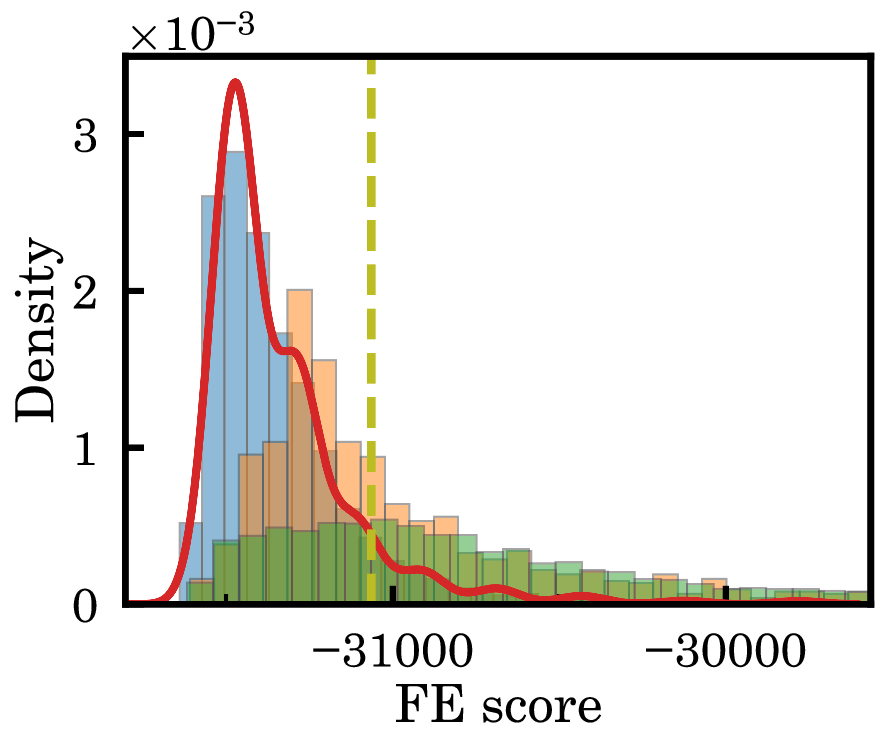} \\
\hspace{0.6cm} (c)
\end{minipage}
\begin{minipage}[t]{0.48\linewidth}
\centering
\vspace{-3.7cm}
\includegraphics[width=0.6\linewidth]{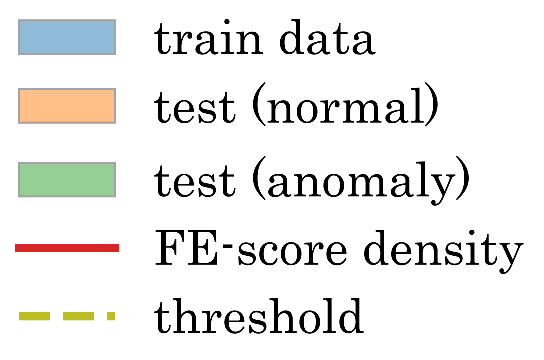}
\end{minipage}
\end{tabular}
\vspace{2mm}
\caption{Normalized histgrams of FE scores for training and test data, FE-score density functions, and thresholds for (a) toy dataset, (b) MNIST, and (c) F-MNIST.
}
\label{fig:energy_Dist}
\end{figure}

The results shown in Fig.~\ref{fig:energy_Dist} demonstrate that the obtained threshold appears to be reasonable. 
In fact, the MCC for the threshold is not significantly lower than the maximum MCC: these values are respectively $0.70$ and $1.00$ for the toy dataset, $0.65$ and $0.74$ for MNIST, and $0.21$ and $0.22$ for F-MNIST.
These results demonstrate that our proposed guideline can be used to set a reasonable threshold.


\section{Conclusion and Future Works} \label{sec:conclusion_future-works}

To facilitate semi-supervised GBRBM-based AD, we designed a measure using the cumulative distribution of the FE score and formulated the measure-based guideline to appropriately set the threshold of the FE score using only normal data. 
The proposed measure is a probabilistic score that expresses the degree of anomaly for given data and is easier to interpret than the FE score.
However, the proposed measure requires an evaluation of the minimum FE score, which is computationally difficult. 
Therefore, we also proposed the evaluation method based on SA, which determines lower FE scores than those obtained using our previously proposed AIS-based method.
Furthermore, the proposed method also yields the data point corresponding to the minimum FE score. 

When setting the FE score threshold using the proposed guideline, it is necessary to specify the anomalous probability $p_\mathrm{anom}$ as a hyperparameter.
The significance level of the proposed measure can be utilized to specify the anomalous probability and might be identified using the FE-score density obtained in this study. 
This task represents a potential future direction of research.
Moreover, RBMs and GBRBMs are often used as models to solve optimization problems~\cite{HatakeyamaSato2022,PhysRevX.11.031034,PROBST2017368}. 
The proposed evaluation method for the minimum FE score optimizes the visible variable corresponding to the data, and can be directly applied to optimization problems. 
This application will be also addressed in subsequent research.

\subsubsection{Acknowledgements} This work was partially supported by JSPS KAKENHI (Grant No. 21K11778) and Innovative Soft Matter Program in Doctoral Course, Yamagata University.

%
%
%
\bibliographystyle{splncs04}
\bibliography{mybibliography}

\begin{thebibliography}{10}
\providecommand{\url}[1]{\texttt{#1}}
\providecommand{\urlprefix}{URL }
\providecommand{\doi}[1]{https://doi.org/#1}

\bibitem{bishop2006pattern}
Bishop, C.M.: Pattern recognition and machine learning. Springer (2006)

\bibitem{cho2011}
Cho, K., Ilin, A., Raiko, T.: Improved learning of gaussian-bernoulli restricted boltzmann machines. In: Artificial Neural Networks and Machine Learning -- ICANN 2011. pp. 10--17. Springer Berlin Heidelberg, Berlin, Heidelberg (2011)

\bibitem{desjardins2010parallel}
Desjardins, G., Courville, A., Bengio, Y., Vincent, P., Delalleau, O., et~al.: Parallel tempering for training of restricted boltzmann machines. In: Proceedings of the thirteenth international conference on artificial intelligence and statistics. pp. 145--152. MIT Press Cambridge, MA (2010)

\bibitem{Do2018}
Do, K., Tran, T., Venkatesh, S.: Energy-based anomaly detection for mixed data. Knowledge and Information Systems  \textbf{57}(2),  413--435 (2018)

\bibitem{fiore2013}
Fiore, U., Palmieri, F., Castiglione, A., De~Santis, A.: Network anomaly detection with the restricted boltzmann machine. Neurocomputing  \textbf{122},  13--23 (2013)

\bibitem{fissore2019}
Fissore, G., Decelle, A., Furtlehner, C., Han, Y.: Robust multi-output learning with highly incomplete data via restricted boltzmann machines. arXiv preprint arXiv:1912.09382  (2019)

\bibitem{Geman1984-wh}
Geman, S., Geman, D.: Stochastic relaxation, gibbs distributions, and the bayesian restoration of images. IEEE Trans. Pattern Anal. Mach. Intell.  \textbf{6}(6),  721--741 (1984)

\bibitem{Xavier2010understanding}
Glorot, X., Bengio, Y.: Understanding the difficulty of training deep feedforward neural networks. In Proc. of the 13th International Conference on Artificial Intelligence and Statistics  \textbf{9},  249--256 (2010)

\bibitem{HatakeyamaSato2022}
Hatakeyama‐Sato, K., Adachi, H., Umeki, M., Kashikawa, T., Kimura, K., Oyaizu, K.: Automated design of li+‐conducting polymer by quantum‐inspired annealing. Macromolecular Rapid Communications  \textbf{43}(20) (2022)

\bibitem{hinton2002training}
Hinton, G.E.: Training products of experts by minimizing contrastive divergence. Neural computation  \textbf{14}(8),  1771--1800 (2002)

\bibitem{hinton2006}
Hinton, G.E., Salakhutdinov, R.R.: Reducing the dimensionality of data with neural networks. Science  \textbf{313}(5786),  504--507 (2006)

\bibitem{Adam2015}
Kingma, D.P., Ba, L.J.: Adam: A method for stochastic optimization. In Proc. of the 3rd International Conference on Learning Representations pp. 1--13 (2015)

\bibitem{MATTHEWS1975442}
Matthews, B.: Comparison of the predicted and observed secondary structure of t4 phage lysozyme. Biochimica et Biophysica Acta (BBA) - Protein Structure  \textbf{405}(2),  442--451 (1975)

\bibitem{Melko2019}
Melko, R.G., Carleo, G., Carrasquilla, J., Cirac, J.I.: Restricted boltzmann machines in quantum physics. Nature Physics  \textbf{15}(9),  887^^e2^^80^^93892 (2019)

\bibitem{neal2001annealed}
Neal, R.M.: Annealed importance sampling. Statistics and computing  \textbf{11}(2),  125--139 (2001)

\bibitem{PhysRevX.11.031034}
Nomura, Y., Imada, M.: Dirac-type nodal spin liquid revealed by refined quantum many-body solver using neural-network wave function, correlation ratio, and level spectroscopy. Phys. Rev. X  \textbf{11},  031034 (2021)

\bibitem{PROBST2017368}
Probst, M., Rothlauf, F., Grahl, J.: Scalability of using restricted boltzmann machines for combinatorial optimization. European Journal of Operational Research  \textbf{256}(2),  368--383 (2017)

\bibitem{Pumsirirat2018}
Pumsirirat, A., Yan, L.: Credit card fraud detection using deep learning based on auto-encoder and restricted boltzmann machine. International Journal of Advanced Computer Science and Applications  \textbf{9}(1) (2018)

\bibitem{salakhutdinov2010}
Salakhutdinov, R., Larochelle, H.: Efficient learning of deep boltzmann machines. In: Proceedings of the thirteenth international conference on artificial intelligence and statistics. pp. 693--700. JMLR Workshop and Conference Proceedings (2010)

\bibitem{sekimoto2023quasi}
Sekimoto, K., Takahashi, C., Yasuda, M.: Quasi-free energy evaluation of restricted boltzmann machine for anomaly detection. IEICE Proceedings Series  \textbf{76}(A3L-44) (2023)

\bibitem{sekimoto2023effective}
Sekimoto, K., Yasuda, M.: Effective learning algorithm for restricted boltzmann machines via spatial monte carlo integration. Nonlinear Theory and Its Applications, IEICE  \textbf{14}(2),  228--241 (2023)

\bibitem{seo2016improvement}
Seo, S., Park, S., Kim, J.: Improvement of network intrusion detection accuracy by using restricted boltzmann machine. In: 2016 8th International Conference on Computational Intelligence and Communication Networks (CICN). pp. 413--417. IEEE (2016)

\bibitem{smolensky1986}
Smolensky, P.: Information processing in dynamical systems: foundations of harmony theory. Parallel distributed processing: Explorations in the microstructure of cognition  \textbf{1},  194--281 (1986)

\bibitem{yasuda2022free}
Yasuda, M., Takahashi, C.: Free energy evaluation using marginalized annealed importance sampling. Phys. Rev. E  \textbf{106},  024127 (2022)

\bibitem{muneki2023new}
Yasuda, M., Xiong, Z.: New learning algorithm of gaussian--bernoulli restricted boltzmann machine and its application in feature extraction. IEICE Proceedings Series  \textbf{76}(A3L-42) (2023)

\end{thebibliography}

\end{document}